\documentclass{article}
\usepackage{spconf,amsmath,graphicx}
\usepackage{amsmath,amssymb,amsfonts}
\usepackage{booktabs}
\usepackage{rotating}
\usepackage{multirow}
\usepackage{adjustbox}
\usepackage{subfigure}
\usepackage{algorithm}
\usepackage{algpseudocode}
\usepackage{array}
\usepackage{multirow}
\usepackage{fancyhdr}
\newcommand{\sd}[1]{$^{\pm \text{#1}}$}

\newcommand{\erot}{image enhanced rotation prediction}
\newcommand{\etal}{et al.}
\newcommand{\TODO}[1]{}
\renewcommand{\TODO}[1]{{\color{cyan} [TODO: {#1}]}}
\newcommand{\FIXME}[1]{}
\renewcommand{\FIXME}[1]{{\color{magenta} [FIXME: {#1}]}}
\makeatletter
\newcommand{\figcaption}[1]{\def\@captype{figure}\caption{#1}}
\newcommand{\tblcaption}[1]{\def\@captype{table}\caption{#1}}
\makeatother

\title{Image Enhanced Rotation Prediction for Self-Supervised Learning}
\name{Shin'ya Yamaguchi$^{\star}$ \qquad Sekitoshi Kanai$^{\star}$ \qquad Tetsuya Shioda$^{\star}$ \qquad Shoichiro Takeda$^{\dagger}$}
\address{$^{\star}$ NTT Software Innovation Center \qquad $^{\dagger}$ NTT Media Intelligence Laboratories}
\begin{document}
\ninept
\maketitle

\thispagestyle{empty}
\renewcommand{\headrulewidth}{0.0pt}
\thispagestyle{fancy}
\chead{\vspace{-2.2cm} \tiny Copyright\copyright 2021 IEEE. Published in the IEEE 2021 International Conference on Image Processing (IEEE ICIP 2021), scheduled for 19-22 September 2021 in Anchorage, Alaska, United States. Personal use of this material is permitted. However, permission to reprint/republish this material for advertising or promotional purposes or for creating new collective works for resale or redistribution to servers or lists, or to reuse any copyrighted component of this work in other works, must be obtained from the IEEE. Contact: Manager, Copyrights and Permissions / IEEE Service Center / 445 Hoes Lane / P.O. Box 1331 / Piscataway, NJ 08855-1331, USA. Telephone: + Intl. 908-562-3966. \vspace{0.3cm}}
\rhead{}
\begin{abstract}
  The rotation prediction (Rotation) is a simple pretext-task for self-supervised learning (SSL), where models learn useful representations for target vision tasks by solving pretext-tasks.
  Although Rotation captures information of object shapes, it hardly captures information of textures.
  To tackle this problem, we introduce a novel pretext-task called image enhanced rotation prediction (IE-Rot) for SSL.
  IE-Rot simultaneously solves Rotation and another pretext-task based on image enhancement (e.g., sharpening and solarizing) while maintaining simplicity.
  Through the simultaneous prediction of rotation and image enhancement, models learn representations to capture the information of not only object shapes but also textures.
  Our experimental results show that IE-Rot models outperform Rotation on various standard benchmarks including ImageNet classification, PASCAL-VOC detection, and COCO detection/segmentation.
\end{abstract}

\section{Introduction}

Self-supervised learning (SSL) is regarded as a promising approach to learn image representations for vision tasks.
To learn useful representations for target vision tasks, SSL uses a pre-training task called {\em pretext-task}.
A pretext-task is a task predicting surrogate supervisions (e.g., rotation degree) defined on unlabeled input images.
Via the pretext-tasks with unlabeled data, convolutional neural network (CNN) models learn representations capturing information in images such as object shapes and textures, which can be helpful for the target task, and achieve higher performance.
To learn more sophisticated representations, various pretext-tasks have been proposed for SSL such as predicting locations of image patches~\cite{doersch_ICCV15_context_prediction,noroozi_ICCV2017_counting_representation,noroozi_CVPR2018_jigsaw++,dosovitskiy_NIPS2014_exemplar}, and predicting differences generated from preprocessing~\cite{zhang_ICCV17_split_brain,zhang_ECCV2016_colorization,gidaris_ICLR18_rotation,feng_CVPR19_rotation_decoupling}.

\begin{figure}[t]
   \centering
   \begin{minipage}{0.3\hsize}
       \includegraphics[width=1.0\columnwidth]{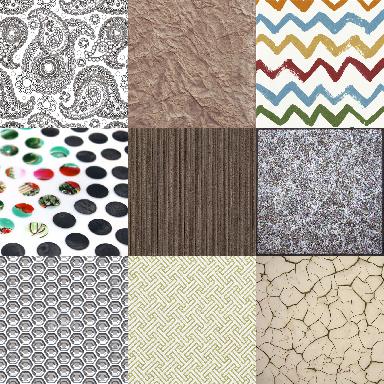}
   \end{minipage}
   \begin{minipage}{0.65\hsize}
       \includegraphics[width=1.0\columnwidth]{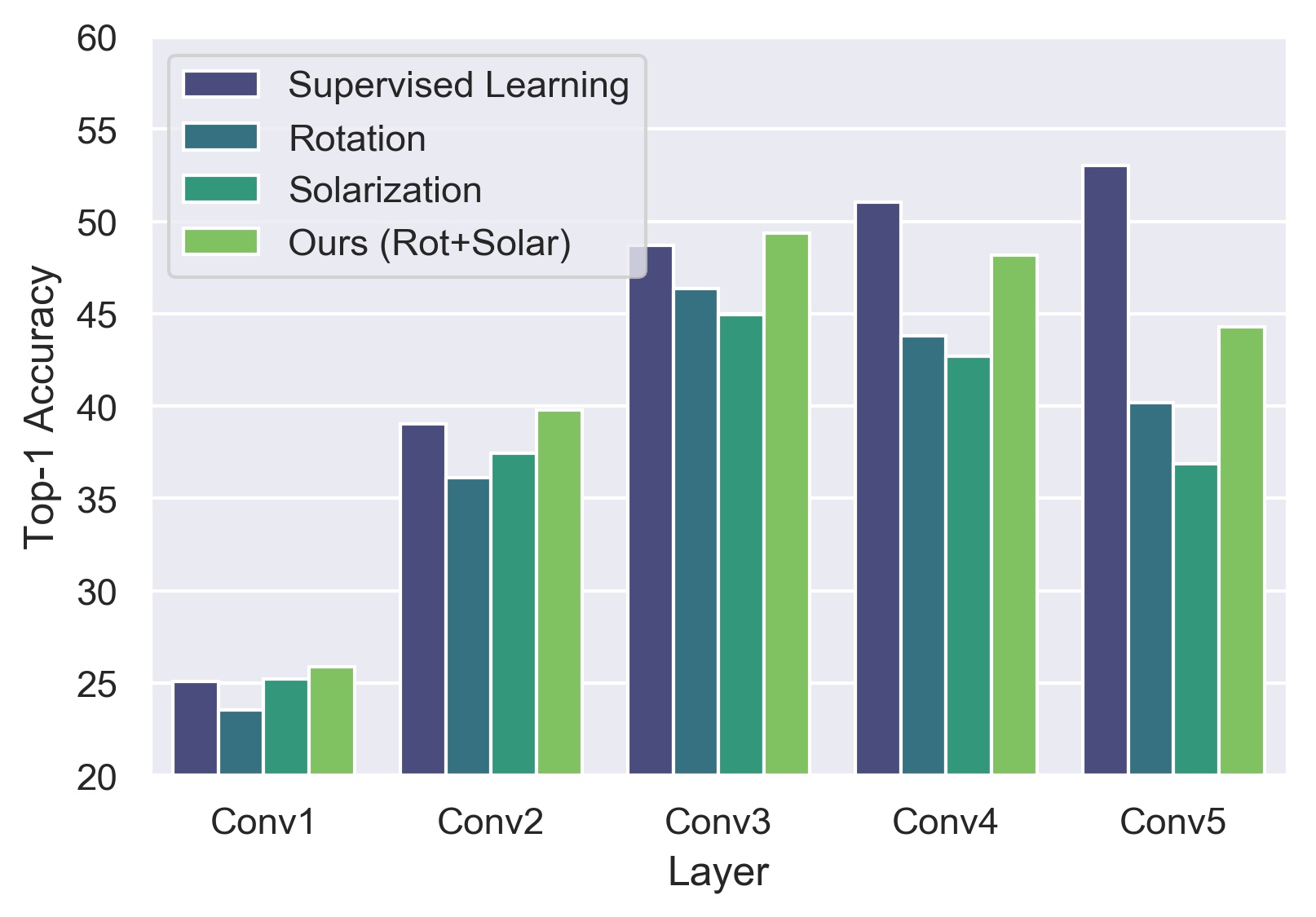}
   \end{minipage}
   \caption{
    The effect of each self-supervised learning (SSL) with ImageNet on DTD classification (AlexNet). Using our image enhanced rotation prediction (Rot~\cite{gidaris_ICLR18_rotation}+Solar) for SSL results in the highest top-1 accuracy; ours is the best for SSL in this task.
   }
   \label{fig:dtd_perf}
 \end{figure}

In such progress of SSL, Gidaris \etal~\cite{gidaris_ICLR18_rotation} have proposed a pretext-task called Rotation that makes a model classify rotation degree of input images (e.g., 0$^\circ$, 90$^\circ$, 180$^\circ$, and 270$^\circ$).
The attraction of Rotation is effectiveness and simplicity compared with other proposed pretext-tasks~\cite{zhang_ICCV17_split_brain,noroozi_CVPR2018_jigsaw++,wu_CVPR18_unsupervised_instance_discrimination}.
By solving Rotation, the models can capture information of object shapes (e.g., the silhouette of cats) because Rotation requires the recognition of the right orientation of objects in input images.
However, Rotation does not consider the existence of information of textures (e.g., black hair of cats) since the rotation transformations only change information of shapes in an image.
For instance, in the case of Describable Texture Dataset (DTD)~\cite{cimpoi14describing_DTD}, which is a dataset for predicting the classes of textures (Fig.~\ref{fig:dtd_perf}:left), models pre-trained with Rotation on ImageNet does not show better target performance (Fig~\ref{fig:dtd_perf}:right).
In general, the information of object shapes plays a crucial role to recognize images~\cite{brendel_ICLR18_approximating_texture,geirhos_2019ICLR_imagenet_texture},
and also the information of textures deeply contributes to the target task performance~\cite{zhang_ICCV17_split_brain,zhang_ECCV2016_colorization,larsson_CVPR17_colorization,lee_ICML20_labelaugmentation}.
Thus, it is desirable to capture information of both object shapes and textures for SSL.

 \begin{figure}[t]
   \centering
       \subfigure[{Input}]{\includegraphics[width=0.19\columnwidth]{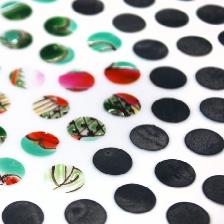}}
       \subfigure[Super.]{\includegraphics[width=0.19\columnwidth]{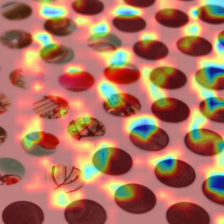}}
       \subfigure[Rotation]{\includegraphics[width=0.19\columnwidth]{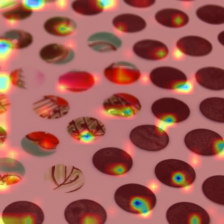}}
       \subfigure[Solar.]{\includegraphics[width=0.19\columnwidth]{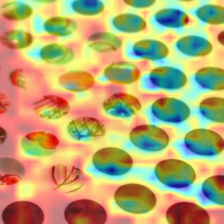}\label{fig:dtd_viz_solarization}}
       \subfigure[Ours]{\includegraphics[width=0.19\columnwidth]{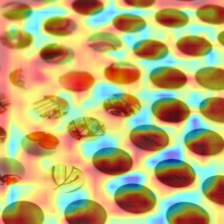}\label{fig:dtd_viz_mit}}
   \caption{
            Attention maps (GradCAM) generated from Conv2 of AlexNet by each pre-training method (same as Fig~\ref{fig:dtd_perf}).
            These heat maps indicate where a trained AlexNet concentrates to recognize an image sampled from DTD.
            The results imply our model can focus on more information of textures.
            Note that Super.\ and Solar.\ stand for Supervised Learning and Solarization, respectively.
            }
   \label{fig:dtd_viz}
 \end{figure}

In this paper, we propose a novel simple method called \erot~(IE-Rot) for SSL, which combines Rotation and image enhancement prediction to learn useful representations for target tasks.
Our key idea is complementarily training a model to capture not only the information of object shapes but also that of textures by leveraging {\em image enhancement} (IE) that modifies appearances of images (e.g., sharpening and solarizing).
In fact, an IE (solarization) prediction for SSL achieves higher accuracy than Rotation when convolution layer 1 and 2 (Conv1 and 2) that captures information of textures~\cite{bau_CVPR17network_dissection} are used for feature extractions (Fig.~\ref{fig:dtd_perf}: right).
IE-Rot generates one transformed image by applying rotation and IE serially, and then solves the classification task for each transformation on this image.
In the training, we optimize each task specific parameter and the feature extractor parameter shared between Rotation and IE prediction.
Finally, we can apply the trained shared parameter for target tasks.
In the example of DTD in Fig.~\ref{fig:dtd_perf}, our models achieve the best accuracy by focusing on textures~(Fig.~\ref{fig:dtd_viz}).
We confirmed that our IE-Rot is superior or comparable to other SSL methods on various target tasks (CIFAR/ImageNet classification, PASCAL-VOC detection, and COCO detection/segmentation). 

\section{Related Work}
Several pretext-tasks for SSL have succeeded to learn useful representations by focusing on various visual information in images.
For instance, the pretext-tasks utilizing image patches are to predict correct patch positions~\cite{doersch_ICCV15_context_prediction,mundhenk_CVPR18_improvements_context} and permutations~\cite{noroozi_ECCV2016_jigsaw},
to measure sum of output logits of image patches~\cite{noroozi_ICCV2017_counting_representation},
and inpainting images masked a partial square region~\cite{pathak_CVPR16_context_inpainting}.
The pretext-tasks of~\cite{zhang_ECCV2016_colorization,zhang_ICCV17_split_brain,larsson_CVPR17_colorization} are to colorize input gray-scaled images by using the output of CNNs.
On the other hand, Rotation~\cite{gidaris_ICLR18_rotation} is a popular method because the model learns powerful representations for various target tasks through solving the simple classification task in advance.
Because of the simplicity, Rotation has been used as a part of the training systems solving tasks in few labeled data settings for classification and image generation task~\cite{lucic_ICML19_s3gan,Chen_CVPR19_Self_Supervised_GAN_with_Rotation,zhai_ICCV19_s4l}.
Additionally, Feng \etal \cite{feng_CVPR19_rotation_decoupling} have presented Decoupling, which is a method enhancing Rotation by adding network architectures for capturing rotation-unrelated information solving instance classification.
In this paper, we aim to improve Rotation while maintaining the simplicity by focusing on the information of textures.
In contrast to Decoupling, our method can be applied without any special modifications for network architectures in Rotation.

\section{Foundation of Rotation}\label{sec:investigation}
In this section, we briefly explain SSL by Rotation.
The pretext-task of Rotation~\cite{gidaris_ICLR18_rotation} is to predict a rotation degree $\tau$ of an input image $x^\tau$ rotated from the original image $x$ to $\tau \in \mathcal{T}$, where $\mathcal{T}=\{1,2,3,4\}$ corresponding to $\{0^{\circ},90^{\circ},180^{\circ},270^{\circ}\}$.
For training, we optimize a network by minimizing softmax-cross entropy loss with respect to the four classes in the set of rotation degrees $\mathcal{T}$.
The objective functions for Rotation are defined as follows:
\begin{equation}\label{eq:obj_rot}
   \min _{\theta, \phi} \frac{1}{N} \textstyle{\sum_{i=1}^{N}} \mathcal{L}\left(x_{i},\theta, \phi \right)
\end{equation}
\begin{equation}\label{eq:loss_rot}
   \mathcal{L}(x_i,\theta, \phi)=-\frac{1}{|\mathcal{T}|} \textstyle{\sum_{\tau \in \mathcal{T}}} \log c\left(F\left(x^\tau_{i};\theta \right);\phi\right)[\tau],
\end{equation}
where, $N$ is a number of training images, $F$ is a feature extractor parameterized by $\theta$
, and $c[\tau]$ is the $\tau$-th element of a classifier $c$ for predicting rotation degrees parameterized by $\phi$.
This is quite easy to implement since we can reuse existing code modules defined for common supervised classification tasks to construct the Rotation loss functions.
In the target task, we use the trained feature extractor $F$ for generating feature maps or initializing target models.

\begin{figure}[t]
  \centering
        \centering
        \subfigure[Rotation~(0$^\circ$, 90$^\circ$, 180$^\circ$, 270$^\circ$)]{\includegraphics[height=1.0cm]{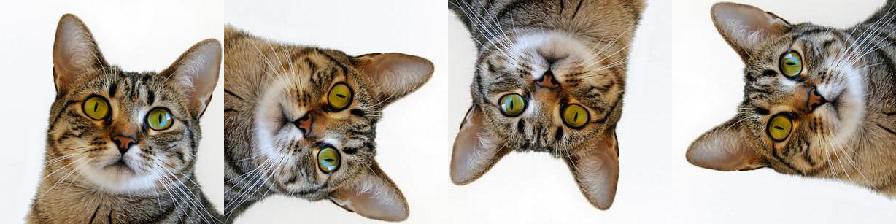}}
        \subfigure[Brightness (0.1, 0.5, 1.0, 1.5)]{\includegraphics[height=1.0cm]{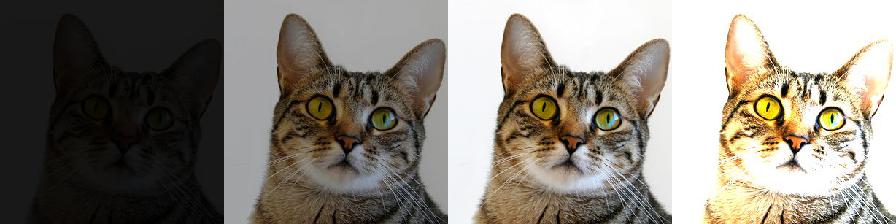}}
        \subfigure[Contrast (0.1, 0.5, 1.0, 1.5)]{\includegraphics[height=1.0cm]{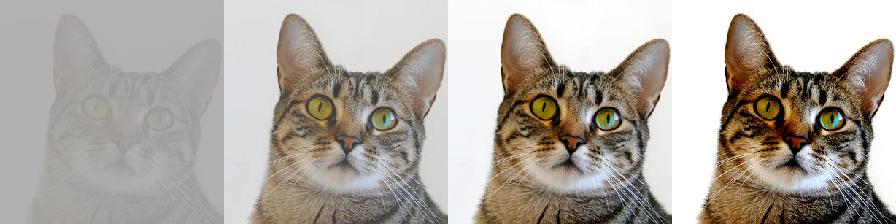}}
        \centering
        \subfigure[Saturation (0.0, 0.5, 1.0, 1.5)]{\includegraphics[height=1.0cm]{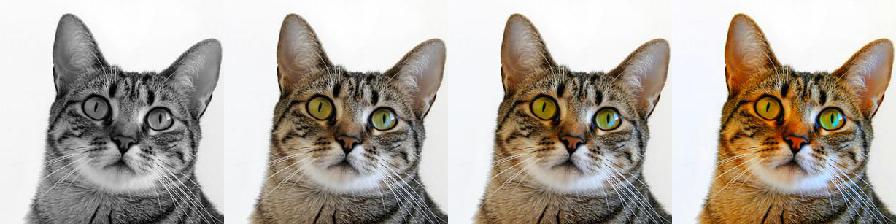}}
        \subfigure[Sharpness (0.0, 0.5, 1.0, 1.5)]{\includegraphics[height=1.0cm]{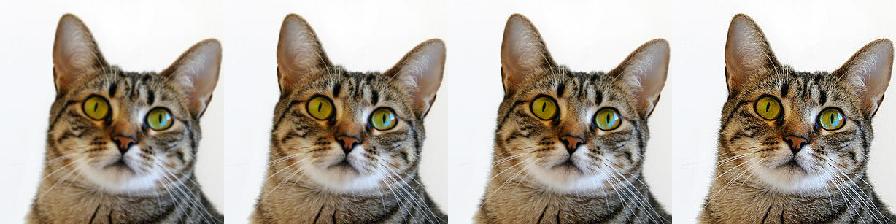}}
        \subfigure[Solarization (0, 85, 170, 256)]{\includegraphics[height=1.0cm]{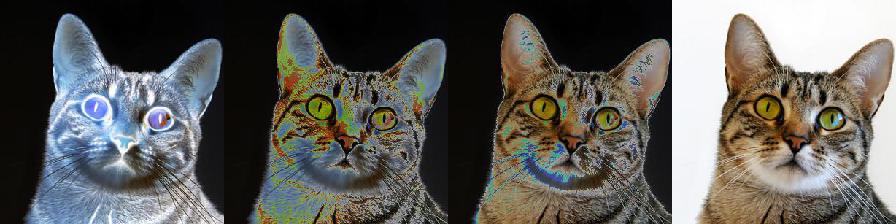}\label{fig:vis_trans_solarize}}
        \label{fig:vis_trans_solarize}
\caption{Samples of the IE transformations designed for IE-Rot. Figures of (b) to (f) describe the transformed images; the left to right order corresponds to the degrees represented as parenthesized values in the captions.}
\label{fig:vis_trans}
\end{figure}

\section{Proposed Method}
\begin{figure}[t]
   \centering
   \includegraphics[width=1.0\columnwidth]{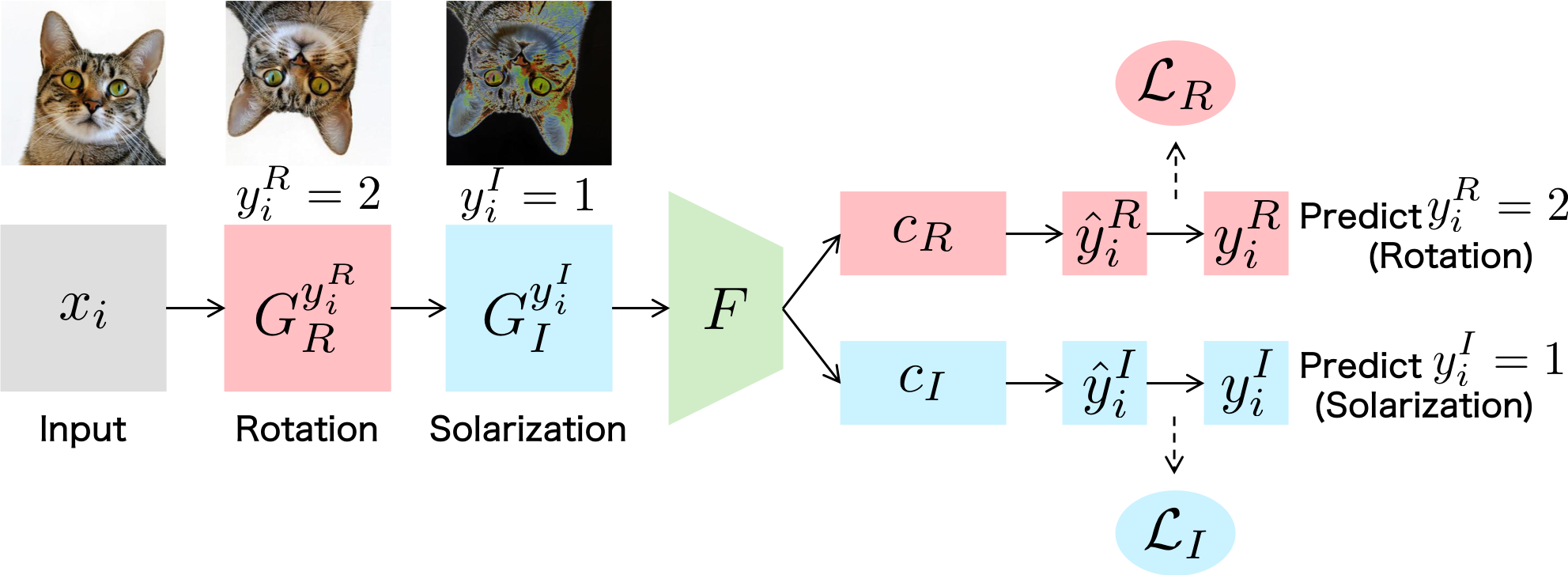}
   \caption{Illustration of IE-Rot in the combination of Rotation and Solarization.}
   \label{fig:proposed}
 \end{figure}

We propose a pretext-task called \erot~(IE-Rot), which combines rotation and IE transformations illustrated in Fig.~\ref{fig:vis_trans}.
IE-Rot solves Rotation and IE prediction by processing one shared image to which two image transformations (i.e., rotation and IE) are applied as shown in Fig.~\ref{fig:proposed}.

\subsection{Image Enhancement Prediction for IE-Rot}
First, we introduce IE prediction as a pretext-task for SSL.
IEs modify information of textures and are often used for data augmentation along with geometric transformations like rotation.
This implies that IEs induce informative differences that are useful for training CNNs.
Furthermore, in contrast to rotation, IEs hardly change geometric information of objects in images; this means IEs and Rotation have little or no interference with each other.
Thus, IEs are suitable for combining with Rotation.

For composing IE prediction, we seek the five well-known IEs: Brightness, Contrast, Saturation, Sharpness, and Solarization.
They are publicly available on open source library such as python image library.\footnote{https://github.com/python-pillow/Pillow}
In the same fashion of Rotation scheme defined as Eq.~(\ref{eq:obj_rot}) and~(\ref{eq:loss_rot}), we arrange the IEs prediction for SSL.
That is, we train a model predicting the discretized degrees of an IE transformation as a classification task.
This formulation is desirable because it can preserve the properties of Rotation requiring no architecture modifications nor specialized loss function.
We set the degree labels to be quartile including the original images as well as Rotation because we have empirically found that the best number is four.
Fig.~\ref{fig:vis_trans} illustrates the transformed images by IEs with arranged degrees.

\subsection{Objective Functions of IE-Rot}
Consider a SSL over an input space ${\cal X}$ and a collection of $d_t$-dimensional task spaces $\{{\cal Y}^t\}_{t\in \{R,I\}}$, where ${\cal Y}^R$ and ${\cal Y}^I$ correspond to Rotation and IE prediction, respectively.
We train a feature extractor $F(\cdot):{\cal X}\rightarrow \mathbb{R}^n$ parameterized by $\theta$ through solving Rotation and IE prediction with each classifier $c_t(\cdot):\mathbb{R}^n\rightarrow\mathbb{R}^{d^t}$ parameterized by the specific parameter $\phi_t$.
We define a set of quartile labels (as discussed in the previous section) for the task $t$ as $Y^t = \{y^{t,1}, y^{t,2}, y^{t,3}, y^{t,4}\}$.
For the training, we use the set of input and task labels: $\{x_i, y^{R}_i, y^{I}_i\}^N_{i=1}$,
where $N$ is the number of input images, $x_i$ is the $i$-th input image in ${\mathcal X}$, and $y^t_i$ is the the label of the task $t$ for $x_i$.
We randomly sample $y^t_i$ from $Y^t$ with uniform distribution for each epoch.
Then, we transform the input image $x_i$ as $x^{\prime}_i = G^{y^R_i}_R \circ G^{y^I_i}_{I}(x_i)$,
where $G^{y^t_i}_{t}$ is the function that returns transformed images according to the given label $y^t_i$ by applying the image transformation corresponding to the task $t$.
Note that a transformed image is generated by applying the two transformations serially, e.g., if we select Rotation and Solarization for the transformations, we first rotate an image, and then, solarize the rotated image.

For the set of transformed images $X^{\prime}=\{x_i^{\prime}\}^{N}_{i=1}$ and the set of corresponding labels for the task $Y_t=\{y^t_i\}^{N}_{i=1}$, a model is optimized by the following formulation:
\begin{equation}\label{eq:obj_mpssl}
   \min _{\theta, \phi_R,\phi_I} \alpha {\mathcal L}_R \left(X^{\prime}, Y_{R}, \theta, \phi_R \right) + (1-\alpha) {\mathcal L}_I \left(X^{\prime}, Y_{I}, \theta, \phi_I \right)
\end{equation}
\begin{equation}\label{eq:loss_mpssl}
   {\mathcal L}_{t}(X^{\prime},Y_t, \theta, \phi_t)= - \frac{1}{N}\!\textstyle{\sum_{i=1}^{N}}\!\log c_t\!\left(F(x^{\prime}_i;\theta);\phi_t\right)\![y^t_i],
\end{equation}
where $\alpha \in [0,1]$ is the task weight balancing the effect of losses across pretext-tasks.
By using the transformed images, we compute a softmax-cross entropy loss by ${\mathcal L}_{t}(\cdot)$ in Eq.~(\ref{eq:loss_mpssl}) as shown in Fig.~\ref{fig:proposed}.
This means that we can easily implement the above loss functions by standard modules equipped in common deep learning frameworks.

\section{Results}
We evaluate the IE-Rot algorithm on various target tasks and datasets.
We confirm (i) the efficacy of IE-Rot on classification tasks, (ii) the performance of IE-Rot against other SSL methods on ImageNet, (iii) the transferability across tasks on PASCAL-VOC and COCO, and (iv) the cause of performance gain of IE-Rot by comparing to data augmentation.

\subsection{Settings}
\paragraph*{Datasets}
We used CIFAR-10/-100~\cite{krizhevsky09_cifar10}, Tiny ImageNet~\cite{wu2017tiny}, ImageNet (ILSVRC 2012)~\cite{russakovsky_IJCV15_imagenet} for classification task, PASCAL-VOC~\cite{Everingham15_IJCV15_PascalVOC}, and COCO~\cite{lin2014microsoft_coco} for detection and segmentation task.
For CIFARs and Tiny ImageNet, we randomly split the train sets into 9:1, and applied the former for training and the latter for validating.
The image size was set to 32$\times$32; in the case of Tiny ImageNet, we randomly cropped 32$\times$32 regions from the original images while training and applied center crop in validation and test time. 
In testing, we used the test sets of CIFAR-10/-100 and the validation set of Tiny ImageNet.
For ImageNet, we randomly split the train set into 99:1, and applied the former for training and the latter for validating.
We set the image size to 224$\times$224 by random crop in training and center crop in testing.
In testing, we used the validation set of ImageNet.
For PASCAL-VOC, we used two types of training sets: the {\tt trainval07} and {\tt trainval07+12} for detection.
We report the evaluation results on {\tt test07}.
For COCO, we trained models on the {\tt train2017} set and evaluated on the {\tt val2017}.

\begin{table}[t]
   \centering
   \caption{
     Comparison of IE-Rot performances among multiple IEs on CIFAR-100. Each cell shows mean test top-1 accuracy of the linear classifier using feature maps generated by the pretrained (frozen) CNN (WRN-40-10, {\tt block-3}).
   }
   \label{tb:cmp_trans}
   \begin{tabular}{lc} \toprule
      Rotation                   & 43.0\sd{0.2}          \\ \midrule
      Rotation+Sharpness     & 44.5\sd{0.4}          \\
      Rotation+Brightness    & 46.0\sd{0.6}          \\
      Rotation+Contrast      & 46.4\sd{0.9}          \\
      Rotation+Saturation    & 46.6\sd{0.1}          \\
      Rotation+Solarization  & {\bf 49.0}\sd{0.3}    \\
      \bottomrule                                
   \end{tabular}
 \end{table}

\paragraph*{Network Architectures}
As the network architectures for the evaluations, we used AlexNet~\cite{Krizhevsky_NIPS12_alexnet}, Wide-ResNet (WRN-40-10)~\cite{Zagoruyko_BMVC16_wide_resnet} and ResNet-50~\cite{He_CVPR16_resnet}.
For the ImageNet experiments, as same the previous works~\cite{zhang_ECCV2016_colorization,gidaris_ICLR18_rotation}, we used a variant of AlexNet architectures, which are modified channel sizes and replaced local response normalization layers to batch normalization layers.
For the PASCAL-VOC and COCO experiments, we used ResNet-50~\cite{He_CVPR16_resnet} as the feature extractor.
For PASCAL-VOC, the model was Faster R-CNN~\cite{ren_NIPS15_fasterrcnn} with a backbone of R50-C4~\cite{he2017maskrcnn}.
For COCO, the model was Mask R-CNN with R50-C4 backbone~\cite{he2017maskrcnn}.

\paragraph*{Training}
We selected settings and parameters for training by reference to~\cite{Zagoruyko_BMVC16_wide_resnet}.
For 32$\times$32 images, we train models by SGD with Nesterov momentum (initial learning rate 0.01, weight decay $5.0\times10^{-4}$, batch size 128).
For ImageNet, we used SGD with Nesterov momentum (initial learning rate 0.005, weight decay $5.0\times10^{-4}$, batch size 256) for the pretext-tasks.
In PASCAL-VOC and COCO evaluations, we used the same setting as the above ImageNet experiments for the pretext-task.
For all pretext-task models, we dropped the learning rate by 0.1 at 30, 60, and 80 epochs and trained models for a total of 100 epochs.
While the original paper of Rotation~\cite{gidaris_ICLR18_rotation} set the total epochs as 30, we found that the larger epochs make both Rotation and IE-Rot achieve better performance.
For determining task weight $\alpha$ in Eq.~\ref{eq:obj_mpssl}, we optimize all IE-Rot models with MGDA-UB which dynamically tunes $\alpha$ for each training step~\cite{Sener_NIPS18_ub_mgda}.

To train target classification models in 32$\times$32 image experiments, we used the logistic regression algorithm in scipy library~\cite{scipy} as the same settings of~\cite{kolesnikov2019revisiting}.
For AlexNet models of target task on ImageNet experiments, we shared the settings with pretext-tasks (see above).
For fine-tuning on PASCAL-VOC and COCO, we run the training and evaluation on detectron2 and used default settings implemented in the detectron2 repository for each task.~\footnote{https://github.com/facebookresearch/detectron2/}
We used {\tt faster\_rcnn\_R\_50\_C4.yaml} for PASCAL-VOC and {\tt mask\_rcnn\_R\_50\_C4\_3x.yaml} for COCO.
We initialized all weight parameters with He normal~\cite{he_ICCV15_initialization}.
In all experiments, each training was run five times with different random seeds, and we report the average scores and the standard deviations.

\begin{table}[t]
  \centering
  \caption{Top-1 linear classification accuracies on ImageNet.
  }
  \label{tb:imagenet_linear_eval}
  \scalebox{0.7}{
  \begin{tabular}{lwc{0.6cm}wc{0.6cm}wc{0.6cm}wc{0.6cm}wc{0.6cm}Wc{0.6cm}} \toprule
      & Conv1 & Conv2 & Conv3 & Conv4 & Conv5   \\ \midrule
     Supervised & 19.3  & 36.3  & 44.2  & 48.3  & 50.5  \\ \midrule
     Random     & 11.6  & 17.1  & 16.9  & 16.3  & 14.1   \\
     Initialization~\cite{krahenbuhl2015data_dependent_init} & 17.5  & 23.0  & 24.5  & 23.2  & 20.6  \\ \midrule
     Inpainting~\cite{pathak_CVPR16_context_inpainting} & 14.1  & 20.7  & 21.0  & 19.8  & 15.5 \\
     Rotation~\cite{gidaris_ICLR18_rotation}& 18.8  & 31.7  & 38.7  & 38.2  & 36.5 \\
     Jigsaw++~\cite{noroozi_CVPR2018_jigsaw++} & 18.9  & 30.5  & 35.7  & 35.4  & 32.2  \\
     InstDisc~\cite{wu_CVPR18_unsupervised_instance_discrimination} & 16.8  & 26.5  & 31.8  & 34.1  & 35.6 \\
     AET~\cite{zhang2019aet} & 19.2  & 32.8  & 40.6  & 39.7  & 37.7 \\
     Decoupling~\cite{feng_CVPR19_rotation_decoupling} & 19.3  & 33.3  & 40.8  & {\bf 41.8}  & {\bf 44.3} \\ \midrule
     Rotation (Our reimpl.) & 21.1  & 34.2  & 39.3  & 38.0  & 36.3  \\
     IE-Rot (Rot+Solar) & {\bf 22.3}  & {\bf 38.4} & {\bf 41.7} & 41.4  & 37.3\\
     \bottomrule                                
  \end{tabular}
  }
\end{table}

\subsection{Comparison of Rotation and IE-Rot}\label{sec:linear_cls_eval}
First, we investigate the most effective IE among five IEs in Fig.~\ref{fig:vis_trans}.
Table~\ref{tb:cmp_trans} shows a summary comparing the performance of IE-Rot and Rotation models on CIFAR-100.
As we expected, all of the IEs succeeded to boost the Rotation performance.
Since the best IE for IE-Rot was Solarization, we report the case of Solarization in the following experiments (represented by Rot+Solar).

\subsection{Evaluations on ImageNet}\label{sec:imagenet_rep_expr}
In order to compare IE-Rot to existing SSL methods, we test IE-Rot on ImageNet representation learning benchmarks following~\cite{zhang_ECCV2016_colorization}.

Table~\ref{tb:imagenet_linear_eval} lists the top-1 classification accuracy of our IE-Rot model on ImageNet using linear classifiers. 
We first trained models with IE-Rot and then trained linear classifier layers for the classification by using feature maps extracted from the trained feature extractor with frozen weights.
The results show that IE-Rot models achieve the best performance by using feature maps on the input-side layers (Conv1-3).
On the other hand, the performance of IE-Rot on the output-side layers (Conv4 and Conv5) did not outperform Decoupling~\cite{feng_CVPR19_rotation_decoupling} that combines Rotation with instance classification in the pretext-task.
This is because the instance classification in Decoupling aims to focus on recognizing object instances; the information of object instances is often considered highly abstract information captured in output-side layers~\cite{bengio2013representation,David_CVPR17_netdissect}.
Although the highly specific information of object instances are useful for the target tasks similar to the pretext-tasks, it can be hard to transfer toward not so related target task.
On the other hand, since IE-Rot models focus on information of textures, the performance boosts appear in the input-side layers that capture information of textures.
Learning such information of textures enhances the transferability of representations that can capture more general information as confirmed in the below experiment.
Note that our reimplementation of Rotation partially improved the performance against the original results reported in~\cite{gidaris_ICLR18_rotation}.
This can be caused by the differences of training epochs between our models (100 epochs) and the original one (30 epochs) reported in~\cite{gidaris_ICLR18_rotation}.

\begin{table}[t]
   \centering
   \caption{
      Fine-tuning results on PASCAL-VOC 2007 and COCO.
   }
   \label{tb:voc_eval}
   \scalebox{0.7}{
   \begin{tabular}{lwc{0.6cm}wc{0.6cm}wc{0.6cm}wc{0.6cm}wc{0.6cm}wc{0.6cm}} \toprule
      \multirow{2}{*}{\underline{PASCAL-VOC}} & \multicolumn{3}{c}{trainval07+12} & \multicolumn{3}{c}{trainval07} \\ 
                         & AP  & AP$_{50}$ & AP$_{75}$ & AP & AP$_{50}$ & AP$_{75}$  \\\midrule
      ImageNet labels    & 52.5 & {\bf 80.3} & 56.2 & 42.2 & {\bf 74.0} & 43.7  \\\midrule
      Rotation           & 51.1 & 76.8 & 55.9 & 45.1 & 70.7 & 48.9  \\
      Decoupling         & 52.8 & 78.1 & 58.6 & 45.4 & 71.0 & 48.7 \\
      IE-Rot (Rot+Solar)    & {\bf 53.6} & 79.3 & {\bf 58.8} & {\bf 45.4} & 71.4 & {\bf 49.0} \\
      \bottomrule
      \multirow{2}{*}{\underline{COCO}} & \multicolumn{3}{c}{Detection} & \multicolumn{3}{c}{Segmentation} \\ 
                         & AP   & AP$_{50}$ & AP$_{75}$ & AP & AP$_{50}$ & AP$_{75}$  \\\midrule
      ImageNet labels    & 39.9 & {\bf 59.2} & 42.5 & 34.5 & 56.5 & 36.1  \\\midrule
      Rotation           & 38.6 & 57.8 & 41.4 & 33.9 & 54.6 & 36.4  \\
      Decoupling         & 38.3 & 57.5 & 41.5 & 33.6 & 54.5 & 35.7 \\
      IE-Rot (Rot+Solar)    & {\bf 39.9} & 58.2 & {\bf 43.5} & {\bf 34.6} & {\bf 56.8} & {\bf 38.1} \\
      \bottomrule                                 
   \end{tabular}}
 \end{table}

\subsection{Task Transferability Evaluations}\label{sec:voc}
In this section, we investigate the transferability of the learned representations across tasks on PASCAL-VOC and COCO datasets.
We used ImageNet as the pretrained dataset and ResNet-50 as the architecture.
We tested IE-Rot, Rotation, and the supervised pretrained models (ImageNet labels) on PASCAL-VOC for detection task and COCO for detection/segmentation tasks.
As another baseline, we reimplemented and tested Decoupling~\cite{feng_CVPR19_rotation_decoupling} methods on the reproducing code provided by the authors.\footnote{https://github.com/philiptheother/FeatureDecoupling}
The linear classification top-1 accuracies on ImageNet were 76.0 of ImageNet labels, 42.2 of Rotation, 57.6 of Decoupling, 51.5 of IE-Rot (Rot+Solar).

Table~\ref{tb:voc_eval} shows the performances of IE-Rot for object detection and instance segmentation task.
We report the results in metrics of COCO-style mean average precision (AP), mean average precision with 0.50 IoU threshold (AP$_{50}$), and mean average precision with 0.75 IoU threshold (AP$_{75}$).
We can see that our IE-Rot succeeds to improve Rotation on both PASCAL-VOC and COCO datasets.
Furthermore, surprisingly, IE-Rot outperforms the models pretrained with ImageNet labels in several cases.
Specifically, IE-Rot performed well on the COCO instance segmentation task that requires models to capture local differences across pixels.
This result implies IE-Rot captures more useful visual information than Decoupling and ImageNet labels as discussed in the previous section.
From these results, we can say that our method captures general information for solving various visual tasks by combining IE prediction with Rotation.

\begin{table}[t]
   \centering
   \caption{
      Comparison of IE-Rot and Data Augmentation (DA). We used WRN-40-10 as the network architecture and Solarization as the IE.
   }
   \label{tb:cmp_da}
   \scalebox{0.8}{
   \begin{tabular}{lccc} \toprule
                              & CIFAR-10           &  CIFAR-100         & TinyImageNet         \\ \midrule
      Rotation                & 74.0\sd{0.5}       & 43.0\sd{0.2}       & 23.4\sd{0.3}  \\ \midrule
      Rotation + DA           & 74.7\sd{0.1}       & 43.3\sd{1.2}       & 23.0\sd{0.4}  \\
      IE-Rot                     & {\bf 75.4}\sd{0.2} & {\bf 49.0}\sd{0.3} & {\bf 26.1}\sd{0.6}  \\
      \bottomrule                                
   \end{tabular}}
\end{table}

\subsection{Comparision to Rotation with Data Augmentation}
Since IEs are used for data augmentation~\cite{Cubuk_CVPR19_autoaug}, improvements of IE-Rot might be caused by implicit augmentation rather than by solving the IE prediction.
Thus, to identify the cause of improvements of IE-Rot, we tested Rotation with data augmentation by IEs and compare it to IE-Rot models.
In the training of Rotation with data augmentation (Rotation+DA), we added the same transformation (solarization) into input images, but trained the models by only Rotation loss.
Table~\ref{tb:cmp_da} shows the comparison of Rotation+DA and IE-Rot models by testing on 32$\times$32 datasets with WRN-40-10.
The performance of Rotation+DA is similar to that of Rotation.
In contrast, IE-Rot outperforms all of the Rotation+DA models so that the improvement of IE-Rot is mainly caused by the simultaneously training.

\section{Conclusion}
This paper presented a novel pretext-task for SSL called \erot~(IE-Rot), which combines Rotation and IEs to learn useful representations focussing on not only information of object shapes but also information of textures.
We confirmed that IE-Rot with Rotation and Solarization improves the target performance across various datasets, tasks, and network architectures.
Although this work focuses on improving Rotation to preserve the simplicity of the pretext-task, the idea of capturing both object shapes and textures can be extended to other pretext-tasks.
As an important future direction, we will apply the idea to contrastive learning such as MoCo~\cite{he_CVPR20_MoCo}, for achieving more powerful representation.

\bibliographystyle{IEEEbib}
\bibliography{icip}

\end{document}